# A Comparative Study on Forecasting of Retail Sales

Md Rashidul Hasan [1], Muntasir A Kabir [2], Rezoan A Shuvro [3], and Pankaz Das [4]

[1,2] University of New Mexico,

[3,4] Marquette University

### Abstract

Predicting product sales of large retail companies is a challenging task considering volatile nature of trends, seasonalities, events as well as unknown factors such as market competitions, change in customer's preferences, or unforeseen events, e.g., COVID-19 outbreak. In this paper, we benchmark forecasting models on historical sales data from Walmart to predict their future sales. We provide a comprehensive theoretical overview and analysis of the state-of-the-art time-series forecasting models. Then, we apply these models on the forecasting challenge dataset (M5 forecasting by Kaggle). Specifically, we use a traditional model, namely, ARIMA ( Autoregressive Integrated Moving Average), and recently developed advanced models e.g., Prophet model developed by Facebook, light gradient boosting machine (LightGBM) model developed by Microsoft and benchmark their performances. Results suggest that ARIMA model outperforms the Facebook Prophet and LightGBM model while the LightGBM model achieves huge computational gain for the large dataset with negligible compromise in the prediction accuracy.

Time-series forecasting; ARIMA; Prophet; LightGBM.

## 1 Introduction

Large retail companies like Walmart, Costco, Amazon, Target, and others have a unique business model where they sell their own products and competitors' products from the same store, either in-store or online. In some cases, their own products compete against a third party product. These companies are a warehouse of massive volume of data, which they store from multiple streams (from transactions, events, inventory to name the few). Companies like Amazon, Walmart help the third party retailers with analytic support from the time series data they acquire from each product and category, as well as gain valuable insight from the data to maximize their business gain. The have a tremendous amount of transactions of multiple products each day. Analyzing such large volume of data and extracting meaningful insight is always a challenge. However, with the advent of cloud computing in recent years analyzing these gigantic data in a small time frame to solve business problems such as predicting future sales and future demand, product recommendation and so forth have become key to success for every company.

As companies scale, forecasting is become an integral and critical part of the business value chain. For example, Walmart can use their historical data to predict the future event sales such as back-to-school, Halloween or Christmas. Moreover, they can use their sales data to have appropriate inventory for perishable products like dairy, bakery and frozen foods. There are certain key aspects to time-series data, namely, trend, seasonality, and noise. For example, during the COVID outbreak in March 2020, people stocked necessary stuffs e.g., bath tissue, which generated significant sales spike in March and eased out in the successive months. If these sudden spike is sales are not identified as noise, models might predict significant demand for bath tissues in the same time next year, which would be catastrophic to business and inventory. Furthermore, certain products show strong seasonalities. For example, Christmas tree sales spike during Christmas not every month of the year. On the other hand, some products might have growth and decline in demand. For example, movie DVD sales are at a decline due to online streaming services and organic produce sales are at an increase. If these trends, either growth or decline, cannot be spotted in time, it might create scarcity (surplus) in inventory in future. Hence, the importance of analyzing product data and the effective translation of the analytics to business has always been a pivotal strength of successful conglomerates, especially in current age.

In this paper, we use time-series data obtained from M5 forecasting- accuracy Kaggle to predict future sales of Walmart products for subsequent 28 days. The dataset was prepared using store sales in three different geographic location in the United States, (Midwest: Wisconsin, West: California, and South: Texas). The data includes item level, department, product categories, and store details Kaggle. The dataset includes 5 years of product sales data including variables such as price, promotions, day of the week, and special events. We classify the models used for our analysis in three types. (1) Statistical models e.g., Autoregressive Integrated Moving Average (ARIMA) Box et al. [2015] (2) gradient boosting based models e.g., light gradient boosting model (LightGBM) Ke et al. [2017], (3) additive models such as Facebook Prophet model Taylor and Letham [2017], which are described below.

ARIMA is the natural extension of Auto regressive and Moving Average models Box et al. [2015]. Along with all the feature and advantages, ARIMA allows to model a non stationary series to a stationary series by taking a sequence of differences. By integrating this differences ARIMA became the more robust to use with any data set and benchmark to compare with other forecasting algorithm. ARIMA essentially can add the difference repeatedly in order to reduce a non-stationary data to stationary. By incorporating with AR order, differencing, and MA order, ARIMA becomes more flexible and robust for forecasting time series data. In addition, by adding the seasonal component to ARIMA model, it becomes an extension of base model that supports the direct modeling with seasonal components.

Here we also apply Facebook prophet which is specially designed to forecast time series data. It has been so far a good competitor to other time series models for several reasons. One, since it is using generalized additive model, it can effortlessly add different variables in the model such as national holidays or events, event types, daily, weekly, biweekly, monthly, quarterly, and yearly trend. In this analysis, we engineered the features of these variable and apply to Facebook prophet model. Two, Facebook prophet can handle missing data and outliers by fitting trends, seasonality, and other dummy variables. This phenomenon of Facebook prophet model provides a significant different from other existing time series models. However, we have applied Facebook prophet model to Walmart sale's dataset with feature engineering and not worrying about any missing data.

Finally, LightGBM works as an out of the box algorithm for the given dataset for two reasons. First, we have found that the one-month sales data we are predicting are not significantly different from the training data consists of nearly 5.3 years. This indicates a natural buying pattern of the Walmart customers, which does not significantly vary on average in the specific item level (here we have 30,490 items). Second, we have created several important features (which are described in Section 5.2) to capture the sales pattern of an item in weekly, biweekly, monthly, bi-monthly, quarterly basis. In addition, we capture special big events (e.g., recreational, religious, cultural, national) with two features of the tree. We emphasize that this feature creation is a novel contribution for this type of sale's dataset, which enables us to use tree-based methods for time series prediction. However, such creation of many features prohibits the use of traditional tree-based algorithms, such as random forest, due to their computation inefficiency with large dataset. Hence, we have used LightGBM to overcome the computation complexity issues.

The paper is organized as follows. A description of the literature is given in Section 2. In Section 3, we explore the data set and introduce the methods for modelling the data in Section 4. We prepare the data for modeling in Section 5 and show the prediction results applying the model in Section 6. We discuss the results and conclude the paper in Sections 7 and 8, respectively.

## 2   Related Works

Forecasting is widely used and studied topics in both academia and industry. Accurate forecasting significantly drives the accuracy of predicting future which is extremely important in the industries such as retail where the uncertainty of future product sales vary abruptly. A comprehensive survey on the time series forecasting, especially prediction methodologies, can be found in Mahalakshmi et al. [2016]. The general approaches described are for time series forecasting are the regression methods (single variable and multiple), stochastic forecasting techniques (e.g., support vector machine (SVM)), soft-computing based forecasting (e.g., artificial neural network (ANN)), fuzzy based forecasting (e.g., fuzzy C-Means (FCM) with ANN). The above methods are used for forecasting electricity loads and prices, trend and seasonality forecasting, stock selection and portfolio construction, etc. Aras et al. compared single method with ensemble technique Aras et al. [2017]. Each of the methods of forecasting has its advantage and can work better with a specific data set, i.e. linear model works better with if the data is linear.

For forecasting time series model, Facebook prophet model so far got a large attention to several researcher from a diverse field of studies. Yenidoğan et al.Yenidoğan et al. [2018] compared the prophet model with ARIMA for forecasting Bitcoin and showed the prophet model outperform ARIMA model. Facebook prophet model has been shown an interesting use in research of environmental phenomena like ground level ambient fine particulate matter ($PM_{2.5}$) concentrations Zhao et al. [2018]. The authors in Zhao et al. [2018] detected weekly and monthly trend of $PM_{2.5}$ concentrations in between 2007 and 2015 for 220 monitoring stations in the United States. Aguilera et al. Aguilera et al. [2019] predicted daily groundwater-level (GWL) for seasonal water management and shown that prophet model outperforms most of relevant method that has been used so far. Scholars are using Long Short-Term Memory (LSTM) for predicting stock market time series model. Predicting stock market prices is always challenging and Facebook prophet does not perform well compare to different neural network's model Mohan et al. [2019]. However, Fang et al. Fang et al. [2019] suggested to use LSTM and prophet to predict the trend and then use the inverse neural network for predicting a time series works better than existing time series models. Another interesting use of this prophet method is predicting an expected meal counts to plan, buy, and produce food for an organization's staffs. Yurtsever and Tecim Yurtsever and Tecim [2020] have shown such application of this model and argued about accuracy and simple use case of this model. The recent pandemic due to Covid-19 virus would have peaked in late October has been predicted by Wang et al.Wang et al. [2020] while using machine learning and Facebook prophet method.

Finally, There have been very little works on predicting time series using tree-based algorithms. This is mostly due to fact that the tree-based algorithms produce regression prediction by assigning average of the leaf (i.e., average of the training data belongs to that leaf) to the data point it is predicting James et al. [2013]. Hence, they cannot capture the trend in the data well, which is very common in the many time series data. To the best of our knowledge, the traditional ARIMA model and tree based Random forest were used for predicting outbreaks of the avian influenza (H5N1) in poultry populations in Egypt Kane et al. [2014]. It has been shown that Random forest achieves better performance compared to the ARIMA model for retrospective and prospective predictions. This is because Random forest was able to capture nonlinear relationship between lagged values and predict values as well as upward shocks of the avian influenza outbreaks. Motivated by this fact, in this paper, we have used LightGBM, which is a tree-based gradient boosting method for Walmart sale's predictions.

# 3 Data Exploration

In this section, we will explore the M5 forecasting (estimate the unit sales of Walmart retail goods) data provided by Walmart, which can be accessed from kaggle Kaggle.

## 3.1 The Dataset

We have used the M5 dataset for this work, which is a grouped time series data of Walmart unit sales across ten stores, located in three different states in the United States (CA, TX, and WI). In this dataset, 3049 products are classified into 3 product categories, which are hobbies, foods, and household, respectively, and 7 different product departments, which can be visualized in figure 1.

## 3.2 Exploratory data analysis

We perform both univariate and multivariate analyses, these analyses also depend on whether the variables are categorical or continuous. Most of the analyses and the findings are based on a series of questions and answers that arise by examining the data.

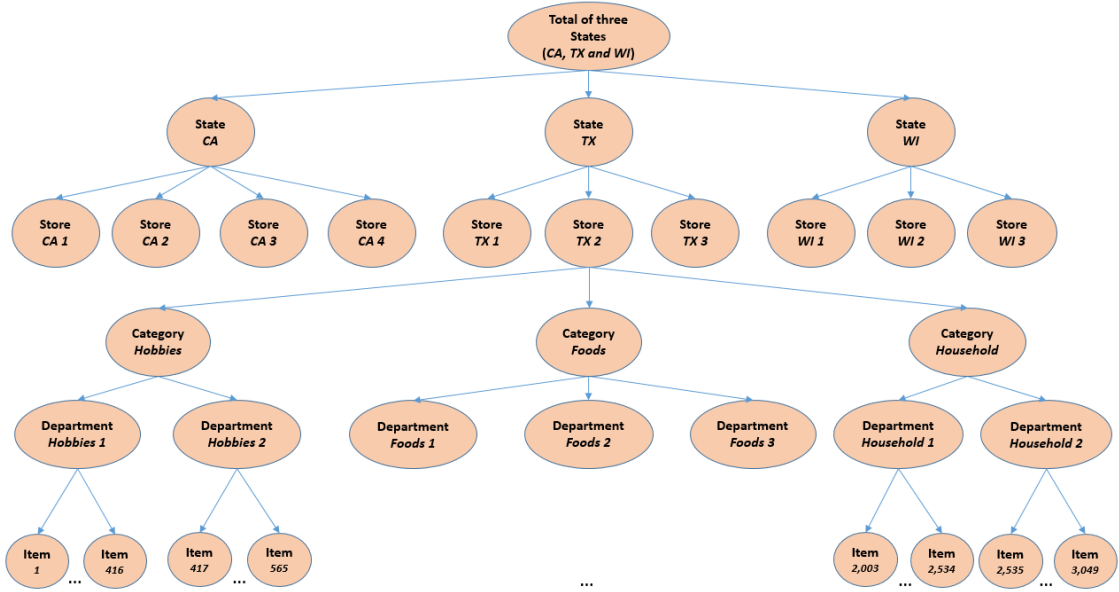

Figure 1: M5-forecasting data categorization Kaggle

### 3.2.1 Time series decomposition

There are two kind of decomposition model: additive and multiplicative. Additive model and Multiplicative modes are described as follows, respectively:

$$y(t) = l(t) + d(t) + s(t) + e(t)$$

, and

$$y(t) = l(t)d(t)s(t)e(t)$$

, where $l(t)$ is the level; $d(t)$ is the trend; $s(t)$ is the seasonality; $e(t)$ is the noise. In a multiplicative time series, the components multiply together to make the time series, i.e., there is an increasing trend, the amplitude of seasonal activity increases. This is appropriate for our sale data since the increase in number of Walmart sales would also increase the seasonal sales.

The three components are shown separately in the bottom three panels of the figure 2. These components can be added together to reconstruct the data shown in the top panel (the original time series). We also noticed that the trend of the data is not strong enough.

### 3.2.2 Influence of events on the product prices

In the figure 3.1, we notice that the average price difference of sell prices between normal days and events days are small. It shows that events do not drive the sell prices. We then calculate the average price of products in normal day, event's day and the price differences between them. We found that during events, the prices are discounted for 20% products and the prices are increased for 26% products. Although the average sell price remains almost the same between normal and event days, on product level, sell prices are driven by the events. Further, investigation of the data reveals that Food_3, Household_2 and Household_1 are the departments with most discounts.

### 3.2.3 Product sales analysis

From figure 3.2, we can see that foods category accounted for most sales; hobbies accounted for least sales. Then, we show the sales in different weekdays. Intuitively, the product sales are higher in weekends compared to weekdays, which can be verified from figure 3.3. Next, we plot the price of product categories over time including the confidence intervals in figure 3.4. We notice that the pricing variation is occurring most in hobbies product. For the food and household category, the prices are negligible in variation compared to hobbies category.

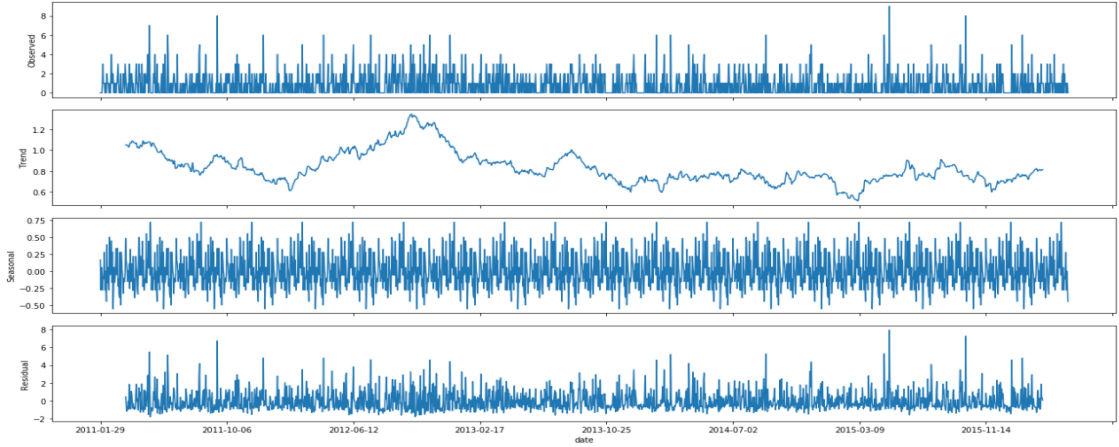

Figure 2: Decomposition of the sample time series data.

## 4   Methods

In this section, we briefly describe the models (i.e., ARIMA, Facebook prophet, LightGBM) that we have used extensively for forecasting unit sales.

### 4.1   ARIMA

The ARIMA modeling procedure was introduced in a pioneering study conducted by Box et al. in 1970 Box et al. [2015]. ARIMA model consider three different component of historical data which is autoregressive terms, moving average, and differencing terms. And very often those components are specified with model like ARIMA(p, d, q) which defines, this model uses p autoregressive terms, q moving average terms and d differences. ARIMA model is based on identifying the structure of the auto-correlations function (ACF) in the data. Classical regression model is often insufficient for explaining all of the interesting dynamics of a time series. For example, an ACF of the residuals of the simple linear regression reveals additional structure in the data that the regression model can not capture Shumway et al. [2000]. Instead, the introduction of correlation as a phenomenon that generates lagged linear relations lead to the development of the autoregressive (AR) and moving average (MA) models. Box and Jenkins Box et al. [2015] added non-stationary part to the mix of AR and MA model, which lead to the widely used ARIMA model.

ARIMA model consists of several theoretical approach that includes an iterative three-step model-building process: model identification, parameter estimation, and diagnostic checking. ARIMA model can be expressed by the following equation:

$$y_t^{'} = c + \phi_1 y_{t-1}^{'} + \cdots + \phi_p y_{t-p}^{'} + \theta_1 e_{t-1} + \cdots + \theta_q e_{t-q} + e_t, \tag{1}$$

where $y_t^{'}$ and $e_t$ are the differenced series, and random error specified at time $t$ respectively. Moreover, $\phi$ and $\theta$ are the parameters of the ARIMA model.

### 4.2   Facebook Prophet

In recent era of e-commerce forecasting, Taylor and Letham Taylor and Letham [2017] introduced a new time series forecasting method based on generalized additive model. This method is well known as Facebook Prophet model for business time series forecasting. The model is used extensively in recent years for data with strong seasonality, and can handle missing data easily.

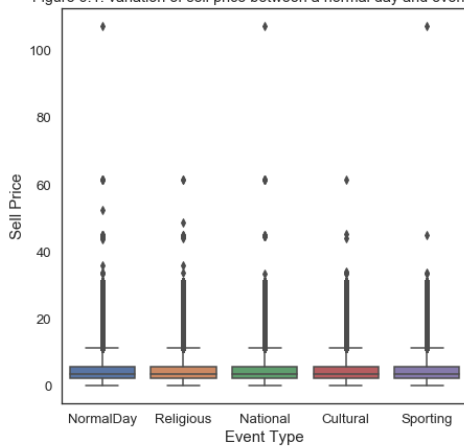

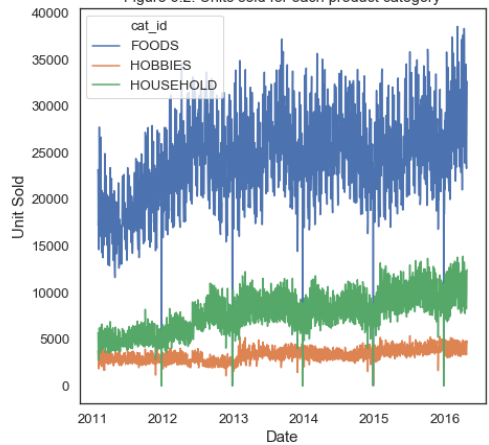

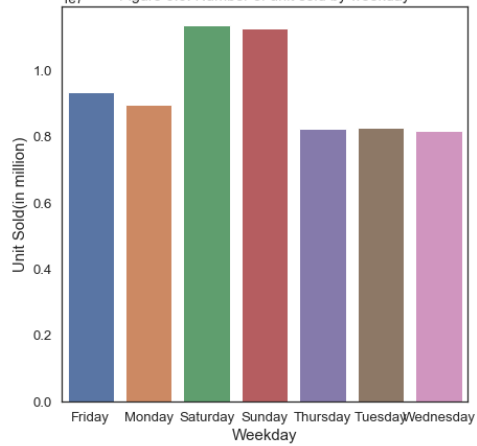

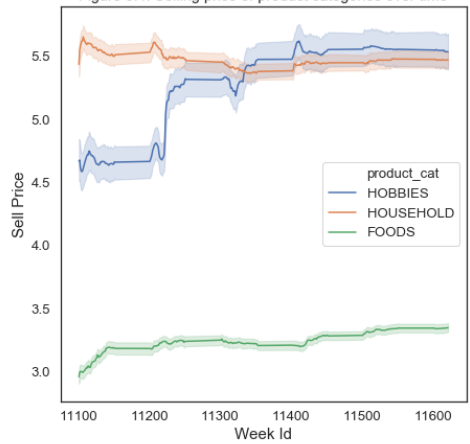

Figure 3: Influence of events on the product sales and prices.

The main model is incorporated with three important components such as trend, seasonality, and holidays. In the simplest form, model can be expressed using the following equation:

$$Y(t) = g(t) + h(t) + s(t) + e(t),\tag{2}$$

where $g(t)$ represents the trend function to model non-periodic effects in the value of the time series; $s(t)$ depicts the seasonality function (weekly, monthly, and yearly); $h(t)$ represents the effects of holidays over the entire year. The error term, $e(t)$ represents the parametric assumption that it is normally distributed.

For forecasting trend, Taylor and Letham Taylor and Letham [2017] have implemented two models: a saturated growth model and a piecewise linear model. Growth trend is typically modeled by logistic growth using the following equation:

$$g(t) = \frac{C}{1 + exp(-k(t - m))},\tag{3}$$

where $C$ is the carrying capacity, $k$ is the growth rate, and $m$ is an offset parameter. Since many business models do not have constant carrying capacity, $C$ can be replaced by $C(t)$. Also, growth rate is not always constant due to new products in the market. Prophet allows $S$ change-points at times $s_j; j = 1, 2, \ldots\ldots$ and can be readjusted the growth rate as $\delta \in R^s$, where $\delta_j$ is the change in rate that occurs at time $s_j$. Now, at time $t$, the rate is $k + \sum_{j:t>s_j} \delta_j$ where $k$ is base rate. This can be easily explained by introducing a vector $a(t) \in (0, 1)^S$ such that

$$a_j(t) = \begin{cases} 1 & , if\ t \geq s_j \\ 0 & , otherwise. \end{cases}\tag{4}$$

Therefore, the rate is $k + a(t)^T\delta$. If the rate $k$ must be adjusted, the offset parameter $m$ also needs to be adjusted to connect the endpoints of the segments. Therefore, correct adjustment at change-point $j$ can be evaluated as $\gamma_j = (s_j - m - \sum_{l<j} \gamma_l)(1 - \frac{k+\sum_{l<j}\gamma_l}{k+\sum_{l\leq j}\gamma_l})$.

Then the logistic model can be defined as piece wise logistic growth model and written as

$$g(t) = \frac{C(t)}{1 + exp(-(k + a(t)^T\delta)(t - (m + a(t)^T\gamma)))}.\tag{5}$$

For linear trend, the model is $g(t) = (k + a(t)^T\delta)t + (m + a(t)^T\gamma)$, where $\gamma_j$ is modified to $-s_j\delta_j$, which makes the function continuous. The change-point $s_j$ can be selected automatically by setting $\delta_j \sim Laplace(0, \tau)$ where $\tau$ has flexibility in controlling the rates.

For seasonality, the regular Fourier series is used, which is

$$S(t) = \sum_{n=1}^{N}(a_n cos(\frac{2\pi nt}{P}) + b_n sin(\frac{2\pi nt}{P})),\tag{6}$$

where $2N$ denotes the numbers of parameters of $a_n$'s and $b_n$'s. They can be chosen from a normal distribution with mean $0$ and standard deviation $\sigma^2$.

Finally, for incorporating holidays into the model, let $D_i$ be the set of past and future dates for $i$th holiday. Then

$$Z(t) = [1(h(t) \in D_1)\ldots\ldots\ldots\ldots 1(h(t) \in D_L)],$$

where $h(t) = Z(t)k;\ \ k \sim N(0, \nu^2)$.

## 4.3   LightGBM

LightGBM is gradient boosting based ensemble decision tree, where the term "light" indicates its faster speed and memory efficient operation compared to conventional gradient boosting techniques such as XGBoost Chen and Guestrin [2016] and pGBRT Tyree et al. [2011]. Hence, LightGBM is very effective in handling large data sets, which we are dealing here. In particular, LightGBM employs two techniques to significantly training time while handling large data sets with negligible compromise on accuracy. Namely, the two techniques are: (1) exclusive feature building (EFB) and (2) gradient-based one side sampling (GOSS). The EFB merges sparse features into one feature using a greedy algorithm to reduce feature space for a tree. On the other hand, the GOSS enables reducing the sample size by using large gradient samples and a fraction of lowest gradient samples which further multiplied by a constant to put more weight on the under-trained data set. In addition, LightGBM also grows tree leaf-wise (grow leaf with maximum error) rather than level-wise to achieve lower loss, which, in turn, helps to achieve higher accuracy with large data sets when handling the overfitting carefully using the depth parameter of the tree. Another notable feature of the LightGBM is that it can handle categorical features without doing one-hot encoding. With all these features LightGBM tends to provide up to 20 times faster training time while achieving similar accuracy compared to conventional gradient boosting decision trees.

In this paper, we employ LightGBM predict the Walmart sales time series for the following 28 days (date range) Kaggle. Here, the use of LightGBM noteworthy due to the large size of the Walmart data sets, which can be easily trained by the LightGBM with a reasonable amount of time using a simple computer. Specifically, an implementation of the LightGBM regression tree with a 4.0 GHz Intel Corei7 CPU (no parallel processing) and system memory 16 GB, which took 39 minutes to train and predict the Walmart sales of 28 days for 30,490 time series.

# 5    Data Preparation

In this section, we briefly describe the data manipulations used to prepare the dataset for applying the various models.

## 5.1    ARIMA/Facebook prophet data preparation

The dataset contains events, sales price, and daily unit sold available in separate files. We combined the files into a aggregated time series file, where entries are timely indexed. First, we reshape the daily unit sales data from wide format to long format so that the item is indexed by each day of sales. After that, we merge the daily unit sales data with calendar data. This enables us to analyze the events information for multivariate analysis and feature engineering. The modified data contains item wise unit sales indexed by date. In addition, for Facebook prophet model, we have added daily, weekly, monthly, quarterly, and yearly trend to capture the seasonality of the data.

## 5.2    LightGBM data preparation

We have created several features to capture the trend of the times series well by the LightGBM algorithm. Table 1 provides a snapshot of the feature-engineered data set that has been used to train the LightGBM algorithm. At first, we create two lags of the of the time series for capturing the weekly and approximately monthly trends, which are denoted by lag_7 and lag_28 in Table 1, respectively. We showed that by adding these two features with the provided features, we can improve the accuracy of the LightGBM. The created daily lags are useful to learn history of the of sales on weekly and monthly basis. However, since these are daily sample values, they often fluctuate abruptly due to outliers and noise and often unable to capture weekly and monthly statistical patterns of the sales. Hence, to capture the average weekly and monthly statistical patterns, we have created rolling means of lag_7 and lag_28 features to include four additional features in the dataset. Namely, rmean_7_7 and rmean_28_7 respectively denote rolling means of 7-days and 28-days of the lag_7 feature. Similarly, rmean_7_28 and rmean_28_28 denote rolling means of 7-days and 28-days of the lag_28 feature. In addition, since this is a time series some other time information is created to add some additional details about sale information. Particularly, week of the year, quarter of year, day of the month are created to incorporate time-based feature in the sales dataset.

### 5.2.1    Parameter optimization for the LightGBM

Table 2 shows the optimized parameter settings for the LightGBM that has been used to predict to Walmart sales. The details about these features can be found in the LightGBM documentation website Microsoft Corporation. The learning rate refers how fast the tree is growing, the smaller the learning rate the slower the learning but the learning accuracy is better given the sufficient number of iterations (num_iterations) to learn. The min_data_in_leaf denotes the minimum data points a leaf requires to have. In addition, the objective of the LightGBM is assumed to Poisson, since the Poisson distribution captures the number of sales, where our target variable is the number of sales (unit sold). In addition, we use root mean squared error (RMSE) between the predicted sales and test sales as an metric to minimize the average error in sale's prediction.

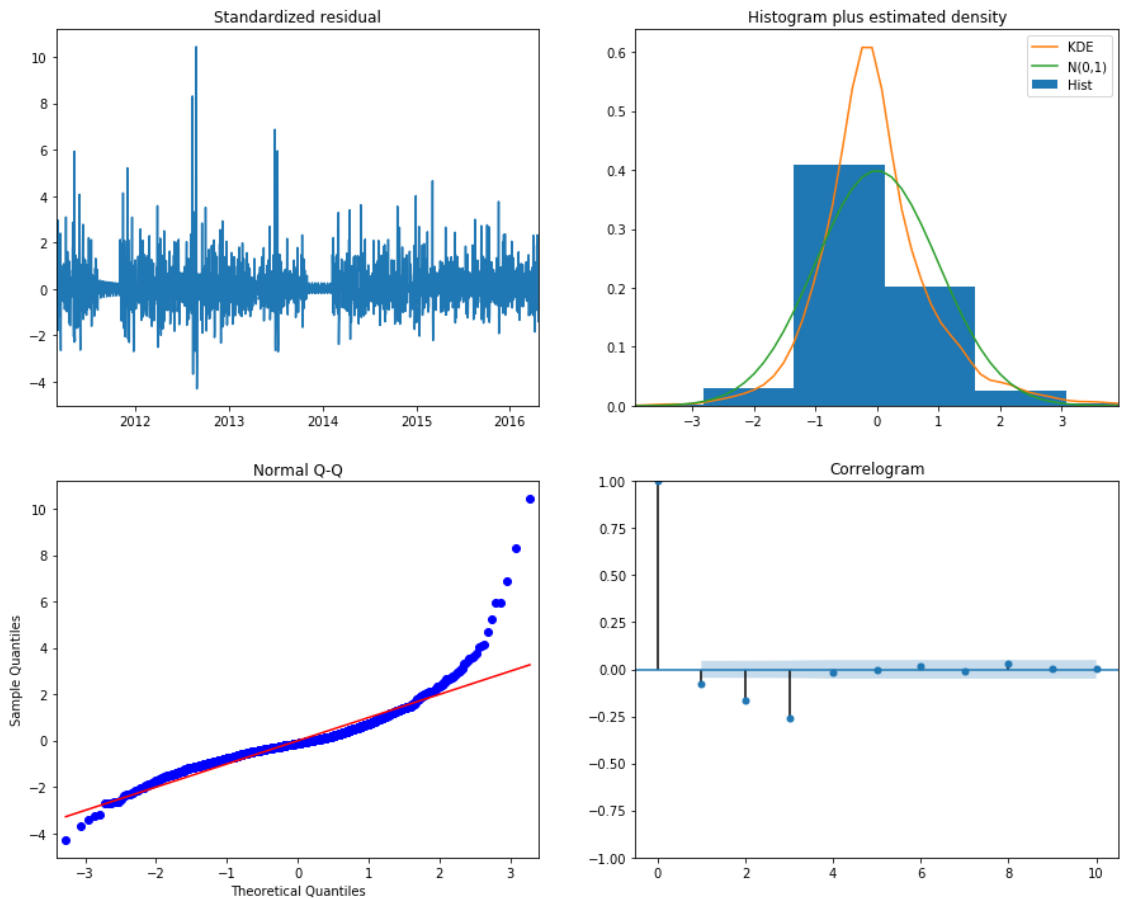

Figure 4: ARIMA model properties

# 6   Results

In this section, we describe the results of our algorithms. First, we apply one product namely HOBBIES_2_120_CA_4_evaluation to create an univariate ARIMA model. This product contains 1913 data points including daily unit sold, event name (e.g. different holidays, sports etc), and event types during 1/29/2011 to 4/24/2016. To choose $p, q, d$ in ARIMA model, we did a grid search of every combinations in between $0$ and $2$ and pick ARIMA(1, 1, 1) model based on the lowest AIC (3618.5482). In figure 4, we plot the ARIMA model properties. We notice from the standardized residual plot (top left of figure 4) that most of the residuals are around zero. However, some outliers also exist, which shows some violation of assumption of normality and constant variance. But there is no obvious trend (top-left figure ). This is also supported by the Q-Q normal and histogram plot (bottom left and top right in figure 4). Based on the correlogram (bottom right) of residuals, it is suggested to use differences to account for the residual correlation.

We also used ARIMA(1,1,1) model for predicting future 28 days unit sold (figure 5). Then, we

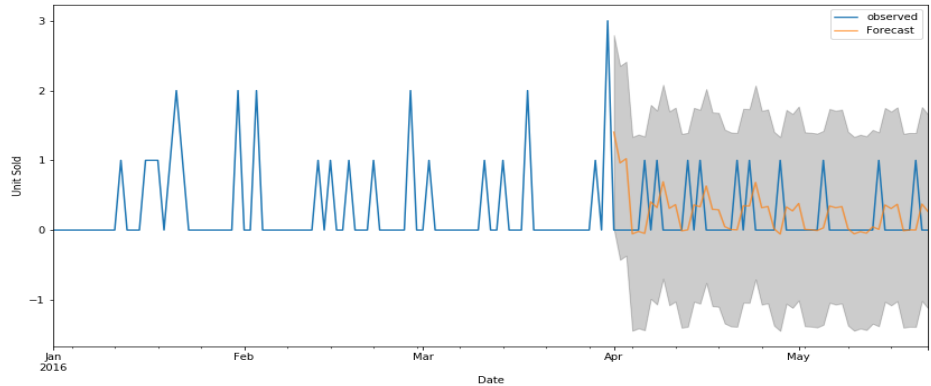

Figure 5: ARIMA model prediction

calculate the Root Mean Squared Error (RMSE) by the following equation

$$RMSE = \sqrt{\frac{\sum_{i=1}^{n}(y_i - \hat{y}_i)^2}{n}},\tag{7}$$

The RMSE for this product is 1.13.

In figure 5, we plot the forecasting for "HOBBIES_2_120_CA_4" for future 57 days started from April 01, 2016 which is colored in yellow along with true or observed value (which is colored in blue) for the same time horizon. We see from the figure that most of the peak sales have been captured by the ARIMA(1,1,1) model.

## 6.1 Prophet forecasting model predictions

For this section, we took one product namely "HOBBIES_1_001_CA_1" as our first step to explain how the Prophet model works. In this product, there are 1913 data points including daily unit sold, event name (e.g. different holidays, sports etc), and event types during 1/29/2011 to 4/24/2016. To analyze the data set, we have added holidays and the trend of daily, weekly, monthly, quarterly, and yearly. The prophet model has intelligence to capture the trend of the data and can predict the future data. The top left plot of Figure 6 shows the trend of the product sales increased in the middle of 2013 and continued to raise afterwards. Holiday trend (top right of Figure 6 ) shows a 600% hike every year on a particular holiday. Middle left of Figure 6 shows the weekly trend and the sale goes up by 40% on Saturday. In the month of March the sales spike to more than 200%, otherwise, sales varies from -200% to 200% throughout the year(middle right of Figure 6). Considering monthly sales, the product unit sold increases by 300% during the month of January(bottom left of Figure 6). Finally, the quarter trend shows some ups and downs of the product sales in between -30% to 30% (bottom right of Figure 6). However, on daily, Weekly, monthly, quarterly, and yearly seasonality show some variations, which lead the model to predict the future sales.

After training with this data set, we predict for future 28 days unit sold. Figure 7 depicts the 28 days predicted the unit sold which has an RMSE of 1.71. We noticed that Facebook prophet gives some negative values ( bottom left corner in Figure 7) and can not capture the higher values (after unit sold number 4 (top right corner) in Figure 7). The negative values are not realistic. That's why we converted it from negative to "zero" meaning nothing has been sold of that particular product on that day.

Table 1: A snapshot of the engineered features that have been used to train the LightGBM algorithm along with original features.

| lag_7 | lag_28 | rmean_7_7 | rmean_28_7 | rmean_7_28 | rmean_28_28 | week | quarter | mday |
|-------|--------|-----------|------------|------------|-------------|------|---------|------|
| 3 | 2 | 1.85 | 1.14 | 1.39 | 1.67 | 10 | 1 | 8 |
| 1 | 0 | 2 | 1 | 1.392 | 1.61 | 10 | 1 | 8 |

Table 2: Optimal parameters for the LightGBM.

| LightGBM parameter name | Value |
|-------------------------|-------|
| objective | Poisson |
| metric | RMSE |
| learning rate | 0.001 |
| num_iterations | 1000 |
| bagging frequency | 1 |
| min_data_in_leaf | 5 |

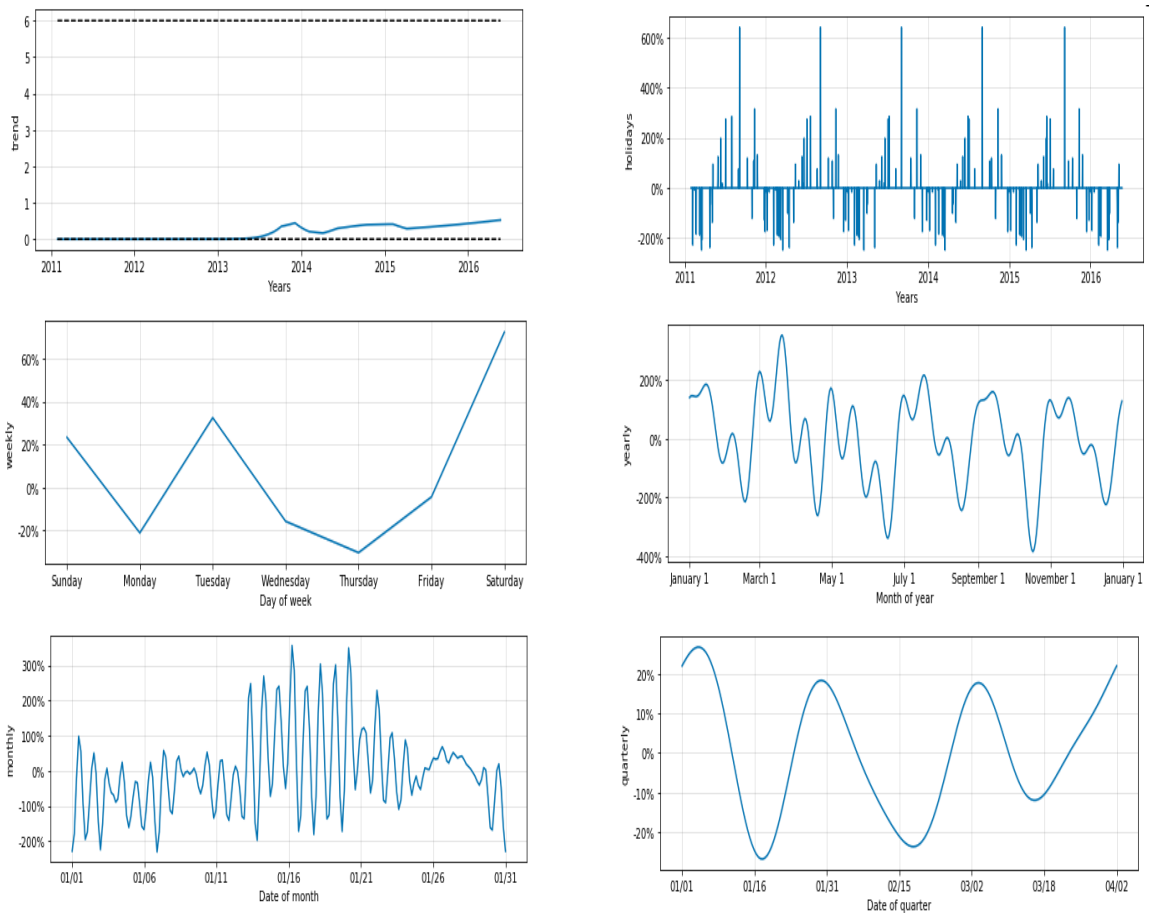

Figure 6: Facebook prophet can easily capture the trend and seasonality of the product's unit sold. Figure shows the trend and seasonality of product "HOBBIES_1_001_CA_1"

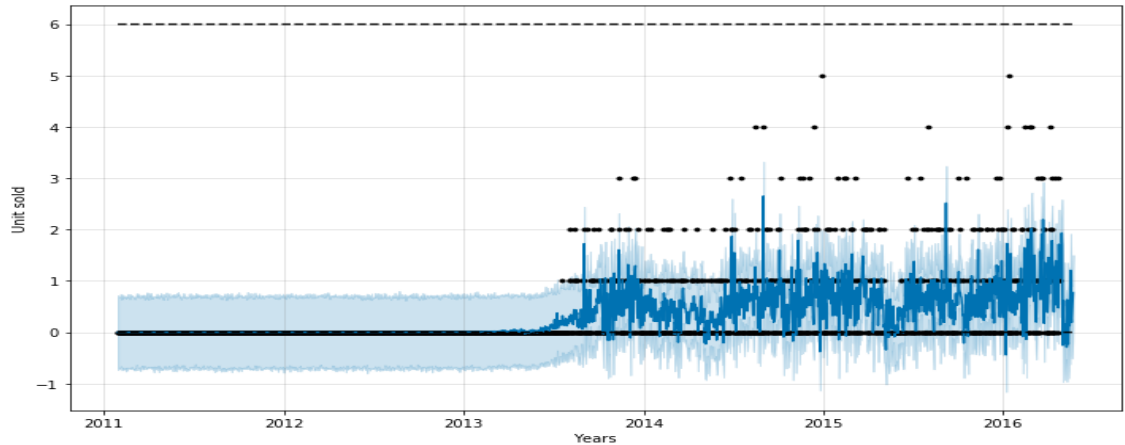

Figure 7: Unit sold for "HOBBIES_1_001_CA_1" are in black dots and Predicted unit sold are in blue line. Prophet model has been instructed to predict future 28 days sold for this product which is also shown in the blue line.

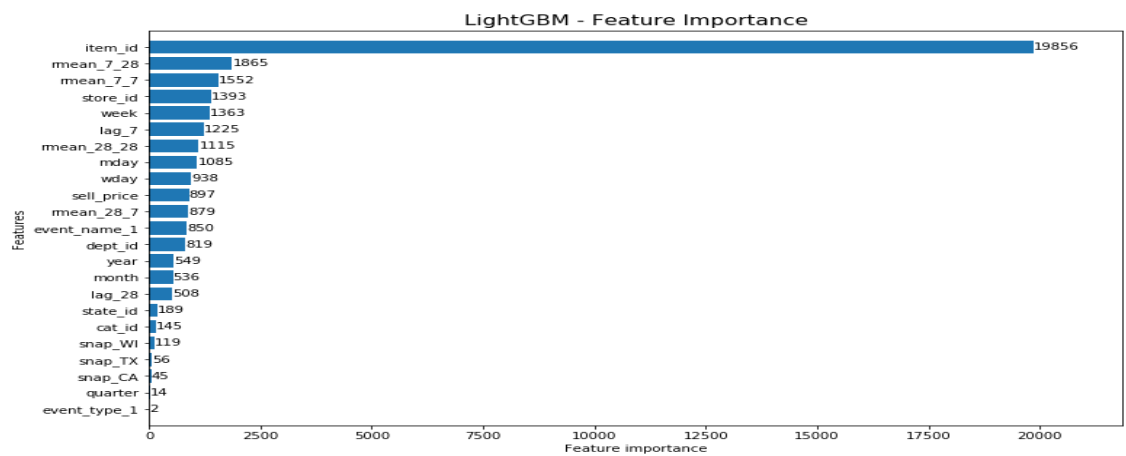

Figure 8: Feature importance of the LightGBM for predicting sales of 30,490 products.

## 6.2 LightGBM predictions

At first, we describe applying the LightGBM on predicting sales of a single product to compare the results with our previous methods (ARIMA, Prophet). We found an RMSE of 0.481 for this product using LightGBM (we skip the figure for in the interest of space and to avoid repetition of the similar figures)

Then we predict the sales of all 30,490 products for the 28 days and found an RMSE of 0.32. Interestingly, we have found that different features have their corresponding importance for predicting the sales. We describe the feature importance of the LightGBM with respect to the number of times the feature is used as a splitting feature while growing the tree. Figure 8 shows feature importance of the LightGBM for predicting sales of 30,490 products. Notable, item ID is intuitive that is most of the times to differentiate each item. But then we notice that rolling means are the second and third most important features, which captures the fact that there are monthly and bi-weekly buying patterns from the customers. In addition, store id reflects the fact of region importance of sales of a products. Subsequently, other features also play role in determining the sale of a product.

Most importantly, the reduction of the RMSE is noteworthy as the LightGBM works very well with large datasets. In addition, with huge reduction of training time that emphasizes the scalability of the LightGBM model for this type of large datasets.

## 7    Discussion

For comparison of the performances, we predict subsequent 28 days sales for 100 products in each category using three methods and reports RMSE as shown in Table 3. Overall, ARIMA gives the lowest RMSE (1.098577) compare to other two models. However, both ARIMA and LightGBM perform better and give almost similar value of RMSE (1.098577 and 1.188333) for all products. On the other hand, Facebook prophet does not perform (overall RMSE of 6.9666) well compare to other competitor methods. In addition, based on individual categories, "Household" provides lowest RMSE (0.83701) while using ARIMA and highest RMSE (11.2851) while using Facebook prophet.

Table 3: Prediction RMSE for each method.

| Forecasting methods | Household | Hobby | Food | Total |
|---|---|---|---|---|
| ARIMA | 0.83701 | 0.96462 | 1.4941 | 1.098577 |
| Facebook Prophet | 11.2851 | 5.8918 | 3.7229 | 6.9666 |
| LightGBM | 0.867 | 0.972 | 1.726 | 1.188333 |

## 8    Conclusions

In this paper, a comparison of the various time-series models and their performance benchmarking on retail sales prediction was done using a large Walmart retail dataset. The dataset was collected from Kaggle. The dataset included five years of sales data (of three categories) for three states of the United States and twenty-eight days of future sales price prediction was used for performance benchmarking. Three forecasting models, ARIMA, Facebook Prophet and LightGBM were used for the benchmarking. Our prediction results indicate that the ARIMA model, although the classical one, gave the best RMSE after tuning while Facebook Prophet resulted with the worst RMSE. However, LightGBM provides nearly similar RMSE as ARIMA in addition to its considerable faster implementation time when considering the whole dataset. It turns out that Facebook prophet overperformed for these data set. Facebook prophet can easily deal with trends and seasonality, but handle higher values badly compare to other two models here. We believe that further feature tuning and hyperparameter tuning will enahnce the model accuracy. Further works include developing an ensemble model using the benchmarked models and optimize the performance to minimize RMSE further.

   All authors contributed equally in formulating the ideas, simulating the models, writing and editing the manuscript.
   no